\pdfoutput=1
\documentclass[11pt]{article}
\usepackage[final]{acl}

\usepackage{times}
\usepackage{latexsym}

\usepackage[T1]{fontenc}
\usepackage[utf8]{inputenc}

\usepackage{microtype}

\usepackage{inconsolata}

\usepackage{graphicx}
\usepackage[inline]{enumitem}
\usepackage{tcolorbox}
\usepackage{xcolor}

\newtcolorbox{lightverbatim}{
  colback=gray!5,   % Lighter gray background
  boxrule=0.5pt,    % Thin border
  arc=0pt,          % Square corners
  outer arc=0pt,
  top=2pt,          % Padding specifications
  bottom=1pt,
  left=2pt,
  right=1pt,
  boxsep=0pt
}

\usepackage{array}
\newcolumntype{C}[1]{>{\centering\arraybackslash}p{#1}}
\newcolumntype{L}[1]{>{\raggedright\arraybackslash}p{#1}}

\title{A Systematic Evaluation of LLM Strategies for Mental Health Text Analysis: Fine-tuning vs. Prompt Engineering vs. RAG}

\author{Arshia Kermani, Veronica Perez-Rosas, Vangelis Metsis  \\
        Department of Computer Science \\ Texas State University \\ 
        San Marcos, TX 78666, USA\\
\small{
   \texttt{\{arshia.kermani, vperezr, vmetsis\}@txstate.edu}
 }
}

\begin{document}
\maketitle

\begin{abstract}
This study presents a systematic comparison of three approaches for the analysis of mental health text using large language models (LLMs): prompt engineering, retrieval augmented generation (RAG), and fine-tuning. Using LLaMA 3, we evaluate these approaches on emotion classification and mental health condition detection tasks across two datasets. Fine-tuning achieves the highest accuracy (91\% for emotion classification, 80\% for mental health conditions) but requires substantial computational resources and large training sets, while prompt engineering and RAG offer more flexible deployment with moderate performance (40-68\% accuracy). Our findings provide practical insights for implementing LLM-based solutions in mental health applications, highlighting the trade-offs between accuracy, computational requirements, and deployment flexibility.
\end{abstract}


% \fi

\section{Introduction}

The increasing prevalence of mental health conditions, coupled with limited access to mental health professionals, has created an urgent need for scalable approaches to mental health assessment and support. Traditional diagnostic methods in this area often rely heavily on clinical interviews and self-reported questionnaires, which can be time-consuming, subject to human bias, and limited in their reach~\cite{10.1186/s40708-022-00180-6}. Recent advances in large language models (LLMs) present promising opportunities to enhance mental health assessment through automated analysis of text-based data.

LLMs have demonstrated remarkable capabilities in understanding and generating human language, with recent models like GPT-4, LLaMA 2~\cite{touvron2023llama2openfoundation}, and their derivatives achieving unprecedented performance across various natural language processing tasks \cite{brown2020language}. In the medical domain specifically, LLMs have shown potential in tasks ranging from clinical decision support to patient education and medical documentation \cite{thirunavukarasu2023large}. However, their application to mental health assessment presents unique challenges due to the nuanced nature of emotional expression and the critical importance of accuracy in clinical contexts.

Previous research has explored various approaches to leverage LLMs for mental health applications. Studies have investigated the use of zero-shot and few-shot prompt strategies for mental health text classification \cite{lamichhane2023evaluation}, achieving moderate success in tasks such as detecting stress and depression. Other work has examined the potential of fine-tuned models for specific mental health tasks \cite{ezerceli2024mental}, demonstrating improved performance through domain adaptation. However, there remains a significant gap in understanding the relative efficacy of different LLM deployment strategies for mental health assessment tasks.

Our study addresses this gap by conducting a systematic comparison of three distinct approaches for mental health text classification: prompt engineering (including both zero-shot and few-shot variants), retrieval augmented generation (RAG), and fine-tuning. We evaluated these approaches using two complementary datasets: the DAIR-AI Emotion dataset, comprising 20,000 tweets labeled with six basic emotions, and the Reddit SuicideWatch and Mental Health Collection (SWMH), which contains 54,412 posts related to various mental health conditions.

This work makes several key contributions to the field:
\begin{enumerate}
    \item We provide the first comprehensive comparison of prompt engineering, RAG, and fine-tuning approaches for mental health text classification, offering insights into their relative strengths and limitations.
    \item We demonstrate the effectiveness of LLaMA 3-based models for mental health assessment tasks, achieving accuracy rates of up to 91\% on emotion classification and 80\% on mental health condition classification through fine-tuning.
    \item We present practical insights into the implementation challenges and resource requirements of each approach, informing future applications in clinical settings.
\end{enumerate}

Our findings have important implications for the development of automated mental health assessment tools, suggesting that while fine-tuning achieves the highest accuracy, both prompt engineering and RAG offer viable alternatives with different trade-offs in terms of computational resources and deployment flexibility. These results contribute to the broader goal of developing reliable, scalable tools to support mental health professionals and improve access to mental health assessment.

In the following sections, we present related work and background (Section~\ref{sec:background}), detail our methodology (Section~\ref{sec:methodology}), present our experimental results (Section~\ref{sec:results}), discuss their limitations (Section~\ref{sec:limitations}), and conclude in (Section~\ref{sec:conclusion}).

\maketitle

\section{Related Work}
\label{sec:background}

The intersection of large language models (LLMs) and mental health assessment represents a rapidly evolving field with significant potential for improving healthcare delivery. This section examines the current state of LLMs in healthcare applications and their specific developments in mental health contexts.

\subsection{Large Language Models in Healthcare}

Recent advances in LLMs have transformed their potential applications in healthcare \cite{he2023survey}. These models have demonstrated capabilities ranging from clinical decision support and medical documentation to patient education and healthcare communication \cite{thirunavukarasu2023large}. The emergence of domain-specific medical LLMs, such as Med-PaLM 2 and Clinical-Camel, has further enhanced their utility in healthcare settings by incorporating specialized medical knowledge and terminology \cite{singhal2025toward}.

The use of LLMs in healthcare applications typically follows three main strategies: fine-tuning existing models, prompt engineering, and retrieval-augmented generation (RAG). Fine-tuning has shown particular promise in specialized medical tasks, with models achieving performance comparable to healthcare professionals in diagnostic scenarios \cite{singhal2025toward}. Prompt engineering approaches have shown effectiveness in zero-shot and few-shot learning contexts, allowing flexible deployment without extensive retraining \cite{liu2023pre}. RAG methods have emerged as a promising approach for grounding LLM responses with domain knowledge, thereby reducing hallucination and improving reliability \cite{lewis2020retrieval,Gao2023RetrievalAugmentedGF}.

\subsection{Mental Health Text Analysis}

Mental health assessment presents unique challenges for automated analysis due to the subtle nature of emotional expression and the critical importance of accurate interpretation. Traditional approaches to the analysis of mental health text have relied on rule-based systems and classical machine learning techniques, often struggling to capture the nuanced context necessary for an accurate assessment \cite{kazdin2011conceptualizing}.

Recent work has begun to explore the potential of LLMs for mental health applications. Studies have shown promising results in the detection of signs of depression, anxiety, and suicidal ideation from social media posts~\cite{ma2024integrating}. The evolution of LLM capabilities has particular relevance for mental health research. Recent studies have shown that advanced LLM versions can provide human-level interpretations in qualitative coding tasks \cite{dunivin2024scalable} and achieve accuracy comparable to mental health professionals in certain diagnostic contexts \cite{kim2024large}. 

When applied to qualitative analysis, LLMs have demonstrated the ability to perform various analytical approaches, including thematic analysis, content analysis, and grounded theory, as validated by human experts \cite{xiao2023supporting,rasheed2024can}. This suggests potential for enhancing, rather than replacing, traditional qualitative analysis methods in mental health research. However, these applications pose important challenges, including the need for accurate predictions given the critical nature of mental health assessment, concerns about privacy and data security, and the importance of maintaining therapeutic alliance in clinical settings \cite{byers2023detecting}.

\begin{figure*}
    \centering
    \includegraphics[width=0.7\textwidth]{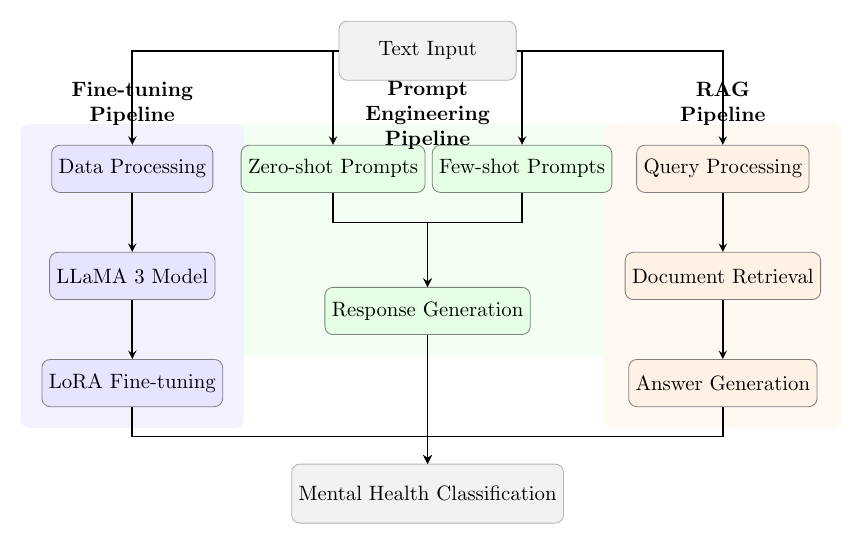}
    \caption{Overview of our experimental framework comparing three LLM deployment approaches (fine-tuning, prompt engineering, and RAG) for mental health text analysis. Each approach processes the same input data through distinct pipelines, enabling systematic comparison of their effectiveness.}
    \label{fig:architecture-overview}
\end{figure*}

\subsection{Evaluation Frameworks}

The evaluation of LLMs in mental health applications requires careful consideration of both model performance and clinical utility. Although traditional metrics such as accuracy and F1 scores provide important quantitative measures, they must be contextualized within the broader requirements of mental health assessment. Recent work has highlighted the importance of developing comprehensive evaluation frameworks that consider not only classification accuracy but also the ability of the model to provide interpretable and clinically relevant output \cite{xu2024mental}.

The existing literature shows particular gaps in understanding the effectiveness of different LLM deployment strategies for mental health applications. Although studies have examined individual approaches, comprehensive comparisons of fine-tuning, prompt engineering, and RAG methods in mental health contexts remain limited. This gap is particularly significant given the practical considerations, such as the availability of resources and the computing requirements involved in deploying these different approaches in clinical settings.

Furthermore, the evaluation of LLMs in mental health applications must consider ethical implications and potential biases~\cite{chancellor2019taxonomy,10.1162/coli_a_00524}. This includes ensuring that models do not perpetuate existing biases in mental health diagnosis and that they maintain appropriate boundaries in therapeutic contexts. These considerations inform both the choice of evaluation metrics and the interpretation of results in mental health applications of LLMs.

\section{Methodology}
\label{sec:methodology}

This study implements and evaluates three distinct approaches for LLMs in mental health text analysis: fine-tuning, prompt engineering, and retrieval-augmented generation (RAG). We utilize the LLaMA 3 model architecture, specifically the 8B parameter version, as our base model across all experiments to ensure fair comparison. Figure~\ref{fig:architecture-overview} presents the overall pipeline of our experimental framework. As shown in the Figure, our prompt engineering approach investigates both zero-shot and few-shot learning capabilities of the LLaMA 3 model for mental health text analysis. %Figure~\ref{fig:Fig1} illustrates our prompt engineering framework and its components.

\subsection{Datasets}

Our evaluation employs two complementary datasets that capture different aspects of mental health and emotional expression in text.

\paragraph{DAIR-AI Emotion Dataset}
The DAIR-AI~\cite{saravia-etal-2018-carer} dataset comprises \texttt{20,000} tweets labeled with one of six fundamental emotions: \textit{joy}, \textit{sadness}, \textit{anger}, \textit{fear}, \textit{love}, or \textit{surprise}. During our experiments, we maintain the original paper's data split: training set: 16,000 samples (80\%); validation set: 2,000 samples (10\%); test set: 2,000 samples (10\%). Table~\ref{tab:dairai-dataset-samples} shows a sample of the DAIR-AI dataset.

\begin{table}[!htbp]
\centering
% \resizebox{\columnwidth}{!}{%
\begin{small}
\begin{tabular}{L{0.75\columnwidth} c}
\hline\hline
\textbf{Sample Text} & \textbf{Label} \\
\hline
\texttt{i can go from feeling so hopeless to so damned hopeful just from being around someone who cares and is awake} & sadness \\
\hline
\texttt{im grabbing a minute to post i feel greedy wrong} & anger \\
\hline
\texttt{i am ever feeling nostalgic about the fireplace i will know that it is still on the property} & love \\
\hline
\texttt{ive been taking or milligrams or times recommended amount and ive fallen asleep a lot faster but i also feel like so funny} & surprise \\
\hline
\texttt{i feel as confused about life as a teenager or as jaded as a year old man} & fear \\
\hline
\texttt{i have been with petronas for years i feel that petronas has performed well and made a huge profit} & joy \\
\hline\hline
\end{tabular}
\end{small}

% }
\caption{Samples and labels from the DAIR-AI Dataset.}
\label{tab:dairai-dataset-samples}
\end{table}

\paragraph{Reddit SuicideWatch and Mental Health Collection (SWMH)}
The SWMH~\cite{ji2021suicidal} dataset contains \texttt{54,412} Reddit posts that discuss various mental health conditions. Each post is labeled with one of the following categories: \textit{depression}, \textit{anxiety}, \textit{bipolar disorder}, or \textit{suicidal ideation}. The dataset is divided by the authors that published it as follows: training set: 34,824 samples (64\%); validation set: 8,706 samples (16\%); test set: 10,882 samples (20\%). Table~\ref{tab:dataset_samples_trimmed} presents a sample of the SWMH dataset.

\subsubsection{Data Preprocessing}
We conduct a preprocessing step on both datasets to ensure data quality and standardization. 1) Removal of URLs, user mentions, and special characters. 2) Standardization of text encoding to UTF-8. 3) Truncation of texts exceeding the model's maximum token limit (2048 tokens). 4) Verification of label consistency and removal of any samples with ambiguous or missing labels.

\begin{table}[!htbp]
\centering
% \resizebox{\columnwidth}{!}{%
\begin{small}
\begin{tabular}{L{0.75\columnwidth} C{0.13\columnwidth}}
\hline\hline
\textbf{Sample Text} & \textbf{Label} \\
\hline
\texttt{Wanting to skip my exam on Saturday because I'm so tired and mentally fried that a few days off might help.} & Anxiety \\
\hline
\texttt{Do other bipolar folks have problems with substance abuse? I've had overdoses and ended up in the ICU, and now I take my meds as prescribed.} & Bipolar \\
\hline
\texttt{Anonymous Entry: plz be nice. I've become a deteriorated husk of a person—hopefully this is my last moment of self-awareness.} & Depression \\
\hline
\texttt{I'm pretty sure my friend is suicidal; he keeps saying self-hating things like ``I'm just a little emo prick.'' What do I do?} & Suicide Watch \\
\hline\hline
\end{tabular}
\end{small}
% }
\caption{Samples and labels from the SWMH Dataset.}
\label{tab:dataset_samples_trimmed}
\end{table}

\subsection{Experimental Setup}

Our experiments are run on A100 GPU with 83.48 GB of RAM and 200 GB of disk space on Google Colab Pro+. %We ensured consistent computational environments across all experiments to maintain reproducibility and fair comparison among approaches. 
%We implemented the fine-tuning process using the
We use the 8B parameter version of LLaMA 3~
\cite{grattafiori2024llama3herdmodels}, applying 4-bit quantization to optimize memory usage while preserving model performance. The base model configuration includes a \texttt{float16} precision for computational efficiency and a LLaMA tokenizer with right-padding and end-of-sequence tokens. During fine-tuning, we used the following hyperparameters: learning rate: 2e-4 with cosine schedule; weight decay: 0.001; batch size: 1 per device; gradient accumulation steps: 8; training epochs: 1; maximum steps: -1.

Classification evaluations are performed using F1 score, precision, and recall as our main metrics.%, and confusion matrices to provide detailed insights into model performance.

\subsection{Fine-tuning}

\begin{figure}
    \centering
    \includegraphics[width=.9\columnwidth]{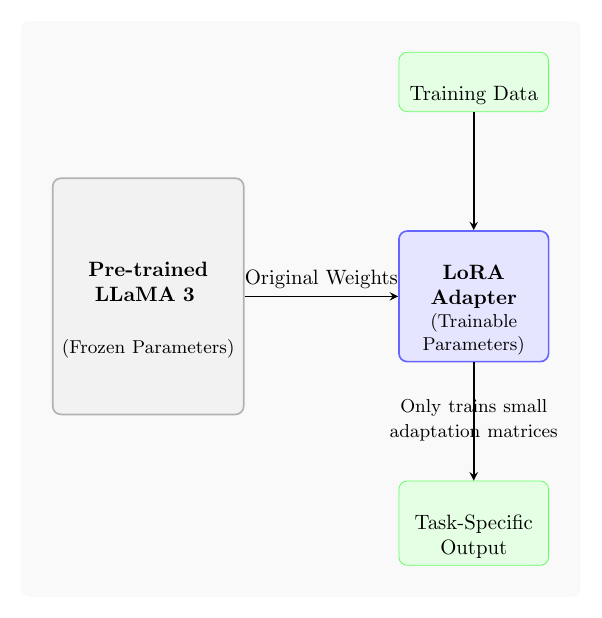}
    \caption{Architecture of our fine-tuning implementation, showing the integration of LoRA for parameter-efficient adaptation, the training process flow, and the evaluation pipeline.}
    \label{fig:Fig2}
\end{figure}

Our fine-tuning approach adapts the LLaMA 3 model to the specific requirements of mental health text analysis while maintaining computational efficiency. Figure~\ref{fig:Fig2} illustrates the fine-tuning architecture and process flow.
%\subsubsection{Model Configuration}
% We implemented the fine-tuning process using the 8B parameter version of LLaMA 3, applying 4-bit quantization to optimize memory usage while preserving model performance. The base model configuration included a \texttt{float16} precision for computational efficiency, and a LLaMA tokenizer with right-padding and end-of-sequence tokens.
%\subsubsection{Parameter-Efficient Fine-Tuning}
To address the computational challenges of fine-tuning large models, we employed Low-Rank Adaptation (LoRA) \cite{hu2021loralowrankadaptationlarge}. This approach significantly reduced the number of trainable parameters while maintaining model performance. Our LoRA configuration included a rank of 64 and an alpha scaling factor of 16.

\subsection{Zero-Shot Prompting}\label{sec:zeroshot}
We use zero-shot prompting to classify text without providing prior examples. %elicit accurate classifications without providing examples, focusing on clear task descriptions and explicit output format requirements. 
% For the DAIR-AI emotion dataset, we use the following prompt. 

We used the following prompt template for both DAIR-AI and SWHM datasets, adjusting for the corresponding labels.

\begin{lightverbatim}
\begin{small}
\begin{verbatim}
Analyze the emotional content in the following 
text and classify it into exactly one of these 
categories: joy, sadness, anger, fear, love, 
or surprise. Provide only the category label as 
output.

Text: [input_text]
\end{verbatim}
\end{small}
\end{lightverbatim}

%For the SWMH dataset, we employed the following zero-shot prompt structure:

%\begin{lightverbatim}
%\begin{small}
%\begin{verbatim}
%Examine the following text and determine which 
%mental health condition is most prominently 
%discussed. Choose exactly one category from: 
%Depression, Anxiety, Bipolar, or SuicideWatch. 
%Return only the category label.

%Text: [input_text]
%\end{verbatim}
%\end{small}
%\end{lightverbatim}

\subsection{Few-Shot Prompting}
In our few-shot approach, we first select \emph{random} examples and their corresponding labels from the training set and provide them as additional input to guide the model's reasoning. We included two examples for each label, ensuring balanced representation across classes. Example prompt:

\begin{lightverbatim}
\begin{small}
\begin{verbatim}
Task: Classify the emotional content of text into 
one of these categories: joy, sadness, anger, fear, 
love, or surprise.

Example 1:
Text: "Finally got my dream job after months of 
trying!"
Emotion: joy

Example 2:
Text: "I miss my old friends so much it hurts."
Emotion: sadness

Example 3:
Text: "How dare they treat people this way!"
Emotion: anger

[Additional examples...]

Now classify this text:
[input_text]
\end{verbatim}
\end{small}
\end{lightverbatim}

% \subsubsection{Prompt Design Considerations}
% We implemented several key design principles across both prompting approaches:

% \begin{enumerate}[nolistsep]
%     \item \textbf{Consistency:} Maintained uniform formatting and instruction structure across all prompts
%     \item \textbf{Clarity:} Provided explicit instructions about output format and category options
%     \item \textbf{Specificity:} Included domain-specific terminology appropriate for each dataset
%     \item \textbf{Conciseness:} Kept prompts focused and free of extraneous information
% \end{enumerate}

% \subsubsection{Implementation Details}
% The prompt engineering pipeline was implemented with specific configurations to ensure optimal performance. The model was used in its base form without additional training, with settings designed to produce deterministic outputs and limit the response length. Automated validation was applied to ensure generated responses adhered to the specified categories.

% \begin{figure}[!]
%     \centering
%     \includegraphics[width=\columnwidth]{figures/prompt-engineering-framework.pdf}
%     \caption{Framework for our prompt engineering implementation, showing the parallel zero-shot and few-shot approaches. The diagram illustrates how different prompting strategies are applied to the same input data, enabling systematic comparison of their effectiveness in mental health text classification tasks.}
%     \label{fig:Fig1}
% \end{figure}

\subsection{Retrieval-Augmented Generation (RAG) }

The success of RAG models hinges on their capability to locate and retrieve pertinent examples and on the LLM's proficiency in effectively using the retrieved information. We believe that this is particularly helpful for the classification of mental health text, where additional context and examples can better inform the model. Since RAG can operate with much fewer training examples than is usually required to fine-tune an LLM model on a specific task, we consider RAG as a middle ground between few-shot prompting and fine-tuning. 

\begin{figure}[ht]
    \centering
    \includegraphics[width=\columnwidth]{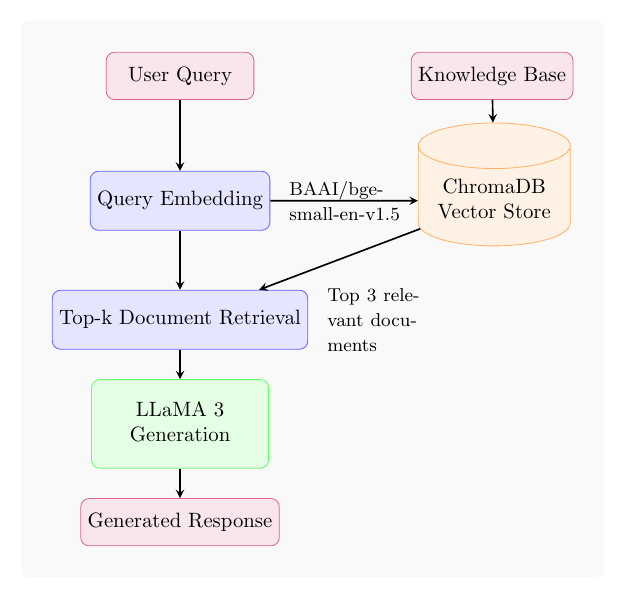}
    \caption{Architecture of our RAG model, illustrating the flow from input processing through retrieval and generation stages. The diagram shows how the system integrates embedded knowledge retrieval with LLM-based classification. % to enhance mental health text analysis accuracy.
    }
    \label{fig:Fig4}
\end{figure}

We implement a RAG model that incorporates relevant contextual information derived from the training dataset during inference time. Figure~\ref{fig:Fig4} presents the overall architecture of our RAG implementation. It retrieves relevant examples from a knowledge base to be used during inference time to inform the model's decision, which are then added as part of the generation input.

\subsubsection{Knowledge Base Construction}
While implementing our model, we constructed a specialized knowledge base to support the retrieval process for each dataset:

\paragraph{Embedding Generation:}
We utilized the BAAI/bge-small-en-v1.5~\cite{bge_embedding} model to generate dense vector representations of training examples. The resulting embeddings are added to a vector database for storage and retrieval.

%The embedding process involved extracting the text component of training prompts and converting them into embeddings using the HuggingFace embedding function. These embeddings are then added to a vector database for storage and retrieval.

\paragraph{Vector Database:}
We use ChromaDB~\cite{chromadb} as our vector database. Our configuration includes cosine similarity as the distance metric, HNSW (Hierarchical Navigable Small World) as the indexing method, and category labels and source information as metadata storage.

    \subsubsection{Retrieval Process}

During retrieval, we start by embedding the input query with the BAAI/bge-small-en-v1.5 model, then we selected the top-k nearest neighbors considering diverse examples across categories, finally, we form an unified context using the retrieved examples.
% \begin{enumerate}[nolistsep]
%     \item \textbf{Query Processing:} Input text is embedded using the same BAAI/bge-small-en-v1.5 model.
%     \item \textbf{Similarity Search:} Top-k nearest neighbors retrieved using cosine similarity.
%     \item \textbf{Diversity Sampling:} Selection of diverse examples across categories.
%     \item \textbf{Context Assembly:} Formation of a unified context document from retrieved examples.
% \end{enumerate}
The retriever returns the three most similar documents for each query, balancing context richness with computational efficiency.

%\subsubsection{Generation Pipeline and Integration Framework}
\paragraph{Generation Component:}
The generation component combines the retrieved context with the input text to produce classification decisions. We use the following prompt template:
\begin{lightverbatim}
\begin{small}
\begin{verbatim}
Review the following examples and context:

[Retrieved Context Documents]

Based on these examples, classify the emotional 
content of the following text into one of these
categories: joy, sadness, anger, fear, love,
or surprise. Provide only the category label.

Text to classify: [input_text]
\end{verbatim}
\end{small}
\end{lightverbatim}

% The RAG pipeline integrates these components through a streamlined workflow:

% \begin{enumerate}[nolistsep]
%     \item Input text preprocessing and embedding generation.
%     \item Relevant examples from training set.
%     \item Context assembly and prompt construction.
%     \item Generation of classification decision.
%     \item Post-processing and validation of output.
% \end{enumerate}

\section{Results}
\label{sec:results}

\begin{figure}[t]
    \centering
    \includegraphics[width=\columnwidth]{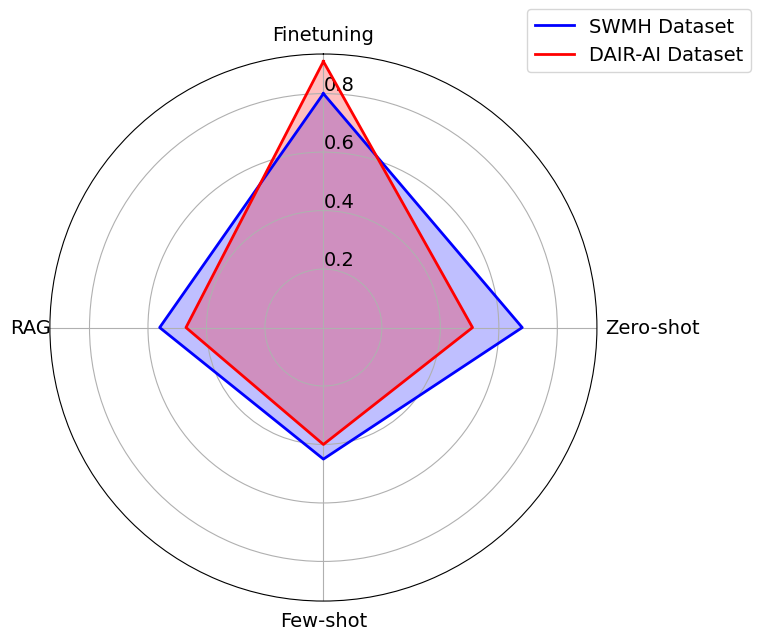}
    \caption{Performance comparison across different approaches for both datasets. The graph shows accuracy scores for fine-tuning, zero-shot prompting, few-shot prompting, and RAG methods.}
    \label{fig:overall-comparison}
\end{figure}

Our experiments show significant performance variations across fine-tuning, prompt engineering, and retrieval augmented generation (RAG) approaches. Figure~\ref{fig:overall-comparison} presents the comparative performance in all methods.

Fine-tuning achieves the best performance, with accuracies of 91\% and 80\% on the DAIR-AI Emotion and SWMH datasets. Notably, zero-shot prompting emerged as the second-best performing approach, reaching 49\% and 68\% on each dataset, surpassing both the few-shot prompting and RAG. This suggests that carefully crafted prompts can effectively leverage the model's pre-trained knowledge for mental health text analysis, even without additional examples or context. However, we should clarify that in this study, we only used simple prompts as in the example shown in section~\ref{sec:zeroshot} and kept them consistent throughout all experiments.

 %The performance disparity between datasets is particularly noteworthy, with the gap between methods being more pronounced in emotion classification compared to mental health condition detection.

\begin{table}[t]
\centering
\begin{small}
\resizebox{\columnwidth}{!}{%
\begin{tabular}{lcccc}
\hline
\hline
\textbf{Method} & \multicolumn{2}{c}{\textbf{DAIR-AI Emotion}} & \multicolumn{2}{c}{\textbf{SWMH}} \\
 & Accuracy &  F1 & Accuracy &  F1 \\
\hline
Fine-tuning & \textbf{91\%} & \textbf{0.87} & \textbf{80\%} & \textbf{0.81} \\
Zero-shot & 49\% & 0.38 & 68\% & 0.67 \\
Few-shot & 39\% & 0.30 & 45\% & 0.57 \\
RAG & 47\% & 0.32 & 56\% & 0.45 \\
\hline
\hline
\end{tabular}
}
\caption{Performance comparison across methods and datasets. F1-scores (macro) provide a balanced measure of performance across all categories.}
\label{tab:performance-comparison}
\end{small}
\end{table}

Table~\ref{tab:performance-comparison} summarizes the classification performance in all methods and datasets. The performance disparity between the different approaches is particularly noteworthy, with the gap being more pronounced in emotion classification compared to the detection of mental health conditions.

Compared to prior work using relation networks \cite{ji2021suicidal}, our approach demonstrates a 15.3\% absolute improvement in classification accuracy and a significant boost in F1-score on the SWMH dataset. While the baseline relied on handcrafted sentiment and topic modeling features for classification, our fine-tuned LLaMA 3 model effectively captures the intricate language nuances in mental health discourse, yielding superior predictive performance. Furthermore, our zero-shot prompting approach (68\% accuracy) surpassed the baseline’s performance, suggesting that LLMs can generalize mental health-related text classification without requiring domain-specific feature engineering.

The greater advantage of the fine-tuning approach in the DAIR-AI Emotion dataset compared to the SWMH can be partially attributed to the dataset sizes (54.4K vs. 20K). A larger training set enables for a more effective fine-tuning, whereas this advantage diminishes and may even be reversed with smaller training sets.

\subsection{Analysis of Best-Performing Methods}

A more detailed evaluation of fine-tuning, zero-shot prompting, few-shot prompting, and RAG reveals distinct patterns in their effectiveness across different classification tasks. Table~\ref{tab:detailed-performance} presents a comprehensive comparison of these approaches in all classification categories.

\begin{table*}[t]
\begin{small}
\centering
\resizebox{\textwidth}{!}{%
\begin{tabular}{llcccccccc}
\hline\hline
\textbf{Dataset} & \textbf{Category} & \multicolumn{2}{c}{\textbf{Fine-tuning}} & \multicolumn{2}{c}{\textbf{Zero-shot}} & \multicolumn{2}{c}{\textbf{Few-shot}} & \multicolumn{2}{c}{\textbf{RAG}} \\
 & & F1 & Prec/Rec & F1 & Prec/Rec & F1 & Prec/Rec & F1 & Prec/Rec \\
\hline\hline
\multicolumn{1}{l}{\textbf{DAIR-AI}} & Joy
   & 0.94 & 0.94/0.93
   & 0.56 & 0.80/0.43
   & 0.35 & 0.35/0.35
   & 0.44 & 0.82/0.30 \\

 & Sadness
   & 0.95 & 0.95/0.94
   & 0.58 & 0.47/0.75
   & 0.24 & 0.20/0.30
   & 0.27 & 0.35/0.22 \\

 & Anger
   & 0.89 & 0.88/0.91
   & 0.46 & 0.39/0.57
   & 0.36 & 0.30/0.45
   & 0.29 & 0.41/0.23 \\

 & Fear
   & 0.89 & 0.89/0.88
   & 0.17 & 0.88/0.09
   & 0.29 & 0.25/0.35
   & 0.31 & 0.45/0.24 \\

 & Love
   & 0.81 & 0.80/0.82
   & 0.25 & 0.26/0.25
   & 0.22 & 0.20/0.25
   & 0.30 & 0.44/0.23 \\

 & Surprise
   & 0.72 & 0.73/0.71
   & 0.26 & 0.24/0.27
   & 0.34 & 0.30/0.40
   & 0.31 & 0.44/0.24 \\

\hline\hline
 & \textbf{Average}
   & 0.87 & 0.86/0.87
   & 0.38 & 0.40/0.36
   & 0.30 & 0.27/0.33
   & 0.32 & 0.45/0.25 \\
\hline
\multicolumn{1}{l}{\textbf{SWMH}} & Depression
   & 0.79 & 0.78/0.80
   & 0.70 & 0.59/0.84
   & 0.52 & 0.74/0.40
   & 0.40 & 0.45/0.36 \\

 & Anxiety
   & 0.86 & 0.87/0.86
   & 0.74 & 0.90/0.63
   & 0.56 & 0.78/0.44
   & 0.61 & 0.72/0.53 \\

 & Bipolar
   & 0.85 & 0.87/0.83
   & 0.62 & 0.88/0.48
   & 0.63 & 0.80/0.53
   & 0.40 & 0.45/0.36 \\

 & Suicide
   & 0.75 & 0.75/0.75
   & 0.61 & 0.68/0.56
   & 0.55 & 0.64/0.48
   & 0.39 & 0.44/0.35 \\

\hline\hline
 & \textbf{Average}
   & 0.81 & 0.82/0.81
   & 0.67 & 0.70/0.64
   & 0.57 & 0.75/0.47
   & 0.45 & 0.52/0.40 \\
\hline\hline
\end{tabular}
}
\caption{Detailed performance metrics for Fine-tuning, Zero-shot, Few-shot, and RAG approaches across all categories. Precision/Recall values are presented as Prec/Rec.}
\label{tab:detailed-performance}
\end{small}
\end{table*}

The fine-tuned model demonstrated varying levels of performance across emotion categories. It achieved exceptional results for basic emotions such as \textit{joy} and \textit{sadness} (F1-scores of 0.94 and 0.95), followed by a strong performance for \textit{anger} and \textit{fear} (both 0.89). More complex emotional states proved more challenging, with \textit{love} achieving an F1-score of 0.81 and \textit{surprise} showing the lowest performance at 0.72. For mental health conditions, the model achieved the highest performance in detecting \textit{anxiety} and \textit{bipolar} disorder (F1-scores of 0.86 and 0.85, respectively) while maintaining robust performance for \textit{depression} detection (F1: 0.79).

Zero-shot prompting showed notably strong performance in mental health condition detection, particularly for \textit{depression} and \textit{anxiety} (F1-scores of 0.70 and 0.74). However, its performance on emotion classification varied considerably. While achieving moderate results for \textit{joy} and \textit{sadness} (F1-scores of 0.56 and 0.58), it struggled significantly with more nuanced emotions like \textit{love} and \textit{surprise} (F1-scores of 0.25 and 0.26). The approach showed particularly low recall for \textit{fear} detection despite high precision, indicating a conservative classification pattern for this category.

\subsection{Analysis of Less Successful Methods}

The evaluation of RAG and few-shot prompting revealed important insights about their practical limitations in mental health text analysis. Table~\ref{tab:alternative-methods} presents the key performance metrics for these approaches.

\begin{table}[t]
\centering
\begin{small}
\resizebox{\columnwidth}{!}{%
\begin{tabular}{lcccc}
\hline
\hline
\textbf{Method} & \multicolumn{2}{c}{\textbf{DAIR-AI}} & \multicolumn{2}{c}{\textbf{SWMH}} \\
& Acc. & Top Category & Acc. & Top Category \\
\hline
RAG & 47\% & Joy (0.44) & 56\% & Anxiety (0.61) \\
Few-shot & 39\% & Anger (0.36) & 45\% & Bipolar (0.63) \\
\hline
\hline
\end{tabular}
}
\caption{Performance summary of RAG and few-shot approaches. The top Category shows the highest F1 score achieved for any single category.}
\label{tab:alternative-methods}
\end{small}
\end{table}

The RAG system achieved moderate performance levels (47\% and 56\% accuracy in DAIR-AI and SWMH, respectively), with effectiveness heavily dependent on retrieval quality. Performance was strongest when highly relevant context was successfully retrieved (64\% accuracy) but dropped significantly with lower-quality retrievals (31\% accuracy). Few-shot prompting showed unexpectedly lower performance compared to zero-shot approaches, suggesting that example-based prompting may introduce conflicting patterns that complicate the classification task in mental health contexts.

Our findings indicate that while RAG and few-shot prompting offer benefits in terms of interpretability and flexibility, their current implementations face significant challenges in achieving reliable performance for mental health text analysis task \cite{chung2023challengeslargelanguagemodels}.

\section{Conclusion}
\label{sec:conclusion}

This study provided a systematic comparison of fine-tuning, prompt engineering, and retrieval augmented generation for mental health text classification. Fine-tuning showed superior performance, achieving 91\% accuracy in emotion classification and 80\% in the detection of mental health conditions, although at the cost of significant computational requirements. Zero-shot prompting emerged as a viable alternative, particularly for mental health condition detection (68\% accuracy), suggesting that carefully designed prompts can effectively leverage pre-trained knowledge when fine-tuning is not feasible. However, both RAG and few-shot prompting showed limited effectiveness, with performance heavily dependent on retrieval quality and example selection.

These findings have important implications for developing automated mental health assessment tools. While fine-tuned models show promise for reliable screening applications, their varying performance across different emotional states and mental health conditions suggests current approaches may be better suited for initial assessment rather than definitive diagnosis. 

Future research directions include the investigation of hybrid approaches that combine the strengths of multiple methods, the development of more efficient fine-tuning techniques, and the exploration of ways to improve the detection of nuanced psychological states. In addition, more work is needed to validate these approaches in clinical settings and across diverse populations.

\section{Ethical Considerations}

This study follows ethical guidelines on data usage, model reliability, and the responsible deployment of large language models (LLMs) for mental health evaluation. The datasets used in this research, DAIR-AI Emotion and SWMH, are publicly available, ensuring transparency and reproducibility. The SWMH dataset consists of publicly shared Reddit posts, while the DAIR-AI Emotion dataset contains labeled social media text. No personally identifiable information (PII) was processed, and no direct engagement with individuals was conducted.

%Given the complexities of using LLMs for mental health classification, particular care was taken to prevent ethical risks such as bias propagation or the generation of harmful content. The models were strictly evaluated under controlled conditions, using predefined inputs that avoided offensive or misleading content. Additionally, no fine-tuning was performed on confidential clinical data or private therapy transcripts, reducing concerns related to data privacy and unintended inferences about mental health conditions.

Automated systems carry the risk of misclassification, especially in sensitive areas such as depression and suicidal ideation. Any potential application of these models outside of research settings would require extensive validation, supervision by clinical professionals, and adherence to ethical and regulatory standards to avoid misinformation or unintended consequences.

\section{Limitations}
\label{sec:limitations}

This study demonstrated the potential of large language models for psychological assessments but also showed a few limitations. Fine-tuning a model as extensive as LLaMA-3 8B required significant computational resources. This dependency on high-end resources limits the accessibility of our approach for researchers with constrained computational capacities. Furthermore, the models were trained and evaluated on the DAIR-AI Emotion and SWMH datasets, which, while diverse, may not fully capture the complexity and variability of real-world psychological text data. This could restrict the generalizability of the findings to other domains, languages, or text formats, e.g., short vs. long text.
%Additionally, the datasets and the pre-trained LLaMA-3 model may carry inherent biases that influence predictions. For instance, mental health classifications derived from Reddit posts may not represent the linguistic patterns found in clinical settings or non-English-speaking populations, reducing model applicability across diverse contexts. Despite leveraging techniques like Low-Rank Adaptation (LoRA) for parameter efficiency, the optimization process required meticulous tuning of hyperparameters and configurations, which may pose challenges for replication without extensive expertise.
%Performance variability was observed in all the methodologies employed. Fine-tuning achieved high accuracy, but demanded substantial computational resources, whereas prompt engineering offered a resource-efficient alternative with moderate accuracy. Retrieval-Augmented Generation (RAG) integrated external knowledge effectively but relied on a well-curated knowledge base, which may not always be available in real-world applications. Ethical concerns, including data privacy and the potential for misclassification in sensitive psychological assessments, remain significant challenges. 
Additionally, our study does not address the practical integration of these tools into clinical workflows, which would require collaboration with domain experts and rigorous validation.

Addressing these limitations in future research could improve the accessibility, generalizability, and ethical applicability of LLM-based psychological assessment tools.

%This research underscores the need for careful consideration when integrating AI into mental health applications. While LLMs provide scalable and efficient tools for analyzing mental health-related text, their role should be complementary rather than definitive. Ethical concerns surrounding privacy, fairness, and potential societal impact must continue to be evaluated to ensure that AI-driven solutions align with best practices in mental health research and care.

\bibliography{refs}

\end{document}